\pdfoutput=1
\documentclass[11pt]{article}

\usepackage[utf8]{inputenc}
\usepackage[T1]{fontenc}
\usepackage{lmodern}
\usepackage[margin=1in]{geometry}
\usepackage{microtype}
\usepackage{graphicx}
\usepackage{booktabs}
\usepackage{tabularx}
\usepackage{array}
\usepackage{amsmath,amssymb}
\usepackage[numbers,sort&compress]{natbib}
\usepackage{titlesec}
\usepackage{caption}
\usepackage{xcolor}
\usepackage{fancyhdr}
\usepackage{enumitem}
\usepackage{hyperref}

\definecolor{linkblue}{HTML}{1F4E79}
\hypersetup{
  colorlinks=true,
  linkcolor=linkblue,
  citecolor=linkblue,
  urlcolor=linkblue,
  pdftitle={Biological-Hybrid Intelligence: A Conceptual Framework for Distributed Biological--Artificial Computation},
  pdfauthor={Michael Taynnan Barros, Sergio Lopez Bernal, Reinhold Scherer}
}

\titleformat{\section}
  {\large\bfseries}
  {\thesection}{0.75em}{}
\titleformat{\subsection}
  {\normalsize\bfseries}
  {\thesubsection}{0.75em}{}
\titlespacing*{\section}{0pt}{2.2ex plus 0.5ex minus 0.2ex}{0.8ex}
\titlespacing*{\subsection}{0pt}{1.7ex plus 0.4ex minus 0.2ex}{0.5ex}

\setlist{nosep}

\newcommand{\safeincludegraphics}[2][]{%
  \IfFileExists{#2}{\includegraphics[#1]{#2}}{%
    \fbox{\parbox[c][42mm][c]{0.88\linewidth}{\centering
      \textbf{Figure file required for submission}\\[2mm]
      \texttt{#2}}}%
  }%
}

\title{\vspace{-1.2em}\bfseries
Biological-Hybrid Intelligence:\\[0.2em]
\Large A Conceptual Framework for Distributed Biological--Artificial Computation}

\author{%
Michael Taynnan Barros\textsuperscript{1,*}\quad
Sergio L\'{o}pez Bernal\textsuperscript{2,*}\quad
Reinhold Scherer\textsuperscript{1}\\[0.8em]
\small\textsuperscript{1}School of Computer Science and Electronic Engineering, University of Essex, Colchester, UK\\
\small\textsuperscript{2}Department of Information and Communications Engineering, University of Murcia, Murcia, Spain\\[0.4em]
\small\textsuperscript{*}Equal contribution\\[0.25em]
\small\href{mailto:m.barros@essex.ac.uk}{m.barros@essex.ac.uk}\quad
\href{mailto:slopez@um.es}{slopez@um.es}\quad
\href{mailto:r.scherer@essex.ac.uk}{r.scherer@essex.ac.uk}\\[0.25em]
\small ORCID: \href{https://orcid.org/0000-0002-9765-7660}{0000-0002-9765-7660};
\href{https://orcid.org/0000-0003-1869-1965}{0000-0003-1869-1965};
\href{https://orcid.org/0000-0003-3407-9709}{0000-0003-3407-9709}
}
\date{}

\begin{document}
\maketitle
\thispagestyle{plain}

\begin{abstract}
Biological and artificial systems offer complementary forms of adaptation, learning, and computation, with advances in in-vitro neurotechnology increasingly enabling bidirectional coupling between them. As these systems become more tightly integrated, a key architectural question is how task-relevant computation should be distributed across both substrates. Yet existing biohybrid solutions optimise the biological substrate, the AI model, or their interface without explicitly addressing how such computation is allocated, reassigned, and evaluated. We introduce Biological-Hybrid Intelligence (BHI), a conceptual framework for distributing computation across adaptive biological and artificial substrates coupled through a bioelectronic interface and coordinated by an orchestrator. BHI treats both substrates as computational entities whose computational responsibilities may change during operation. BHI requires reciprocal co-adaptation and differs from systems that merely decode biological activity, stimulate a living substrate, or adapt a single component. BHI further defines three operating modes: adversarial, collaborative, and codependent, distinguished by whether the substrates compete, divide computational labour, or become mutually necessary for task performance. BHI provides a common basis for comparing computational frameworks, defining benchmarks for latency, viability, interface bandwidth, learning efficiency, and reproducibility. It also highlights governance considerations arising from reciprocal stimulation, adaptation, and data exchange. More broadly, BHI invites computer scientists to consider biological substrates as active computational resources and to ask not only how a task should be computed, but where its computation should reside. BHI therefore reframes biological–artificial integration as a system-level problem of computational allocation, coordination, and control.
\end{abstract}

\noindent\textbf{Keywords:} Biohybrid intelligence; wetware-in-the-loop;
biological computing; neuromorphic interfaces; organoid intelligence;
neural engineering; co-adaptive systems.

\vspace{0.5em}
\hrule
\vspace{0.8em}

% ----------------------------------------------------------
%                   MAIN BODY
% ----------------------------------------------------------
\section{Introduction}

Understanding how intelligent systems can adapt, learn and improve under changing conditions is a long-standing scientific and engineering challenge. This work uses intelligence to mean the capacity of a computational system to learn from experience and improve task performance under changing environmental conditions. Such adaptive problem solving is essential in real-world environments, which are inherently uncertain, dynamic, and non-stationary. Biological nervous systems and artificial intelligence (AI) provide fundamentally different approaches to adaptive computation. Biological systems can support continual adaptation, robustness, and efficient operation under uncertainty, whereas artificial systems provide scalability, reproducibility, explicit memory, planning, and algorithmic flexibility. Because neither substrate is optimal across all conditions, combining these complementary capabilities may enable forms of problem solving that are difficult to achieve with either alone.

Recent advances in in-vitro neurotechnology and AI have enabled biological and artificial computational substrates to interact within closed feedback loops. Systems such as cultured-neuron controllers, neurorobotic platforms, and organoid-based computing demonstrate that both substrates can contribute to task performance through ongoing interaction \cite{kagan2022vitro,li2016robot,cai2023brain,chen2023overview}. As these systems evolve, adaptation is no longer confined to a single substrate. Neural activity, machine learning models, decoding algorithms, stimulation strategies, and control policies may all adapt during operation \cite{george2020plasticity,keren2019biohybrid,barros2025intersection,aboulkheir2026fuzzy}. Consequently, computation becomes distributed across multiple interacting substrates whose contributions may evolve over time. This problem has parallels with multi-agent systems, where complex problem solving requires tasks to be decomposed, allocated among agents with different capabilities, and dynamically coordinated during execution \cite{Skaltsis2021agents}. In biohybrid systems, however, this coordination problem takes a different form because computation is distributed across fundamentally different biological and artificial substrates whose dynamics, adaptation mechanisms, and operating timescales are not interchangeable. The challenge can already be anticipated in systems such as DishBrain or CL1, where cultured neurons learned to improve performance in a Pong environment through closed-loop feedback \cite{kagan2022vitro}. As adaptive AI components become integrated into similar systems, behaviour may emerge from reciprocal adaptation across substrates rather than from any single component \cite{madduri2026computational}. The central question is therefore no longer how individual components learn, but how computation should be allocated, coordinated, and evaluated when multiple adaptive substrates contribute to the same task.

Existing approaches provide methods for modelling neural activity, designing interfaces, and implementing adaptive controllers, but remain largely component-centric. They offer limited guidance for distributing computational functions across biological and artificial substrates, identifying reciprocal adaptation, or evaluating systems whose organisation evolves during operation. As a result, it becomes difficult to explain performance improvements, compare alternative system designs, benchmark progress, or assign responsibility for system behaviour. The gap is therefore primarily architectural rather than technological: the constituent technologies already exist, but a general framework for organising and evaluating distributed biological-artificial computation remains lacking \cite{kagan2022vitro,cai2023brain,basso2026advanced,sono2026online,mullett2026topology,barros2021engineering}. To address this challenge, this article makes four main contributions: (1) we introduce \textit{Biological-Hybrid Intelligence} (BHI), a conceptual framework for organising computation across adaptive biological and artificial substrates, defining the roles of the biological substrate, AI, biohybrid interface and orchestrator, and making computational responsibility across the system explicit; (2) we define a three-condition test for determining when a system qualifies as BHI through reciprocal, co-adaptations; (3) we define three operating modes (adversarial, collaborative, and codependent), that provide distinct patterns for organising biological–artificial computation; and (4) we provide evaluation criteria and a research roadmap for comparing computational frameworks and advancing reproducible BHI systems. Throughout this article, intelligence refers to measurable adaptive task performance rather than consciousness, sentience, or subjective experience.

\section{Biological-Hybrid Intelligence Framework}

We formalise BHI as a computational framework in which task-relevant computation is distributed across coupled biological and artificial substrates rather than assigned to a single substrate. The framework separates task computation from system coordination. Two substrates perform task computation: Living Intelligence (LI), a living neural culture, and AI, whose computational roles depend on the operating mode and on the intrinsic computational characteristics of each substrate. The \emph{Orchestrator} coordinates their interaction through the Biohybrid Interface Layer (BIL) and, when present, a cyber-physical system (CPS). It manages the allocation of computational responsibilities between LI and AI without performing task computation itself, making explicit which substrate performs each task-relevant computation and how that allocation may evolve during adaptation. 

\begin{figure}
  \centering
  \includegraphics[width=0.95\linewidth]{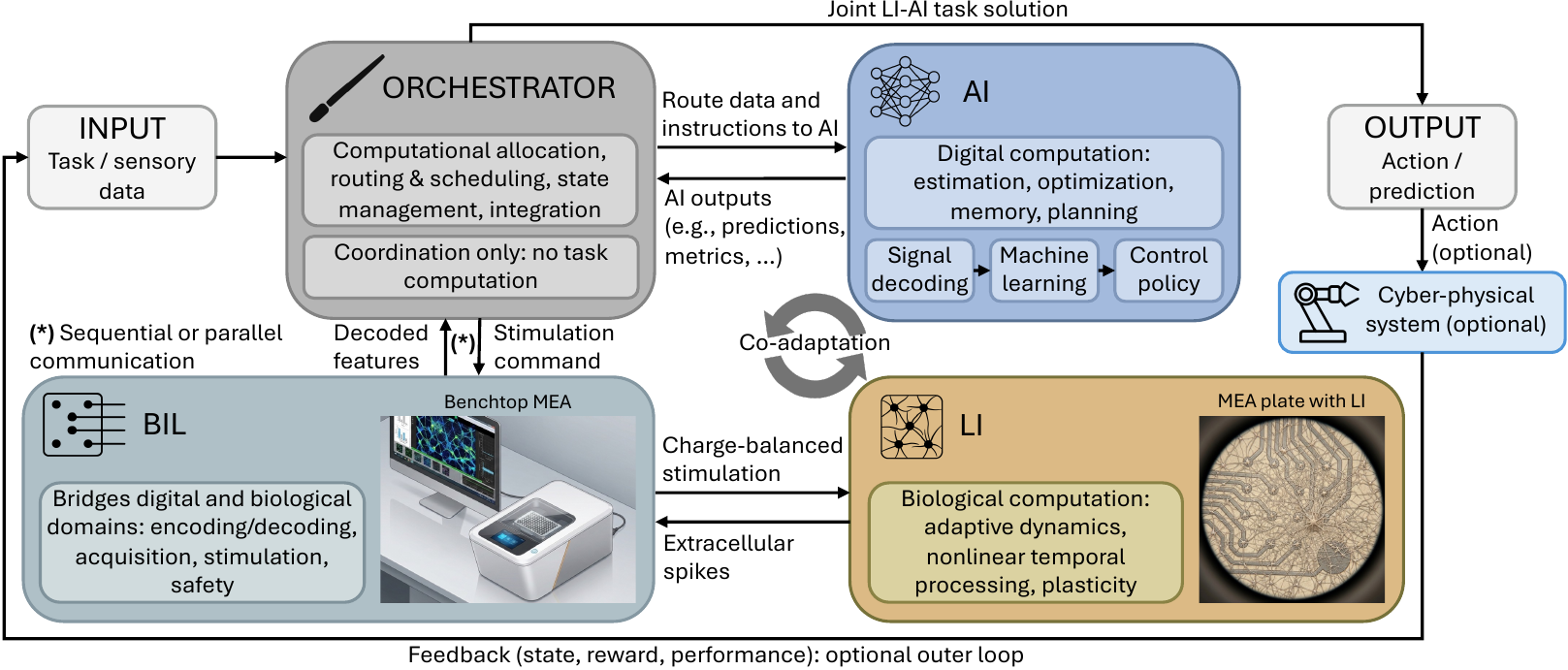}
  \caption{Overview of the proposed framework. For a given task, the Orchestrator assigns computation between AI and LI without performing the computation itself. The BIL connects the two by decoding extracellular activity and converting digital commands into charge-balanced stimulation pulses. AI and LI then contribute according to their capabilities and the selected operating mode. The framework also supports AI–LI co-adaptation and optional interaction with a CPS through an outer feedback loop.}
  \label{fig:scenario}
\end{figure}

Figure~\ref{fig:scenario} presents the proposed framework and illustrates the interaction between its main components. Given an input task, the Orchestrator coordinates execution by allocating computational responsibilities, routing information, scheduling exchanges, maintaining interaction state and integrating the contributions of AI and LI. The BIL acquires biological activity from LI, decodes it into digital features, and translates stimulation commands into charge-balanced stimulation. AI processes routed data and instructions and returns computational outputs, while LI contributes through its adaptive biological dynamics. Their coordinated contributions produce a joint LI–AI task solution, which can generate an action or prediction. This output may drive a CPS, whose resulting state, reward, or performance feedback can be returned through the optional outer loop to support subsequent interaction and adaptation.

Distributed computation alone is insufficient for BHI. The biological and artificial substrates must also influence and adapt to one another within the same closed loop. Neural activity must update the AI's model or policy, while AI-generated stimulation or embodied feedback must influence the biological substrate, producing measurable, interaction-dependent adaptation over time in both. This requirement for reciprocal adaptation distinguishes BHI from systems based solely on neural decoding, open-loop stimulation, or closed-loop systems with adaptation confined to a single substrate. Whether such coupling can produce system-level capabilities unavailable to either substrate alone remains a hypothesis to be tested. Sustained co-adaptation also requires the interaction to remain compatible with the temporal, physiological and interface constraints of the biological substrate. 

\subsection{What Qualifies as BHI}
A system qualifies as BHI only when three conditions are met within the same closed-loop interaction: (1) activity in the biological substrate modifies a learning-relevant state of the AI, (2) the AI influences the biological substrate through stimulation or embodied feedback, and (3) both systems undergo measurable, interaction-dependent adaptation over time. The adaptation required for BHI occurs within AI and LI; the Orchestrator enables their exchange through routing and synchronization but is not itself adaptive. Systems exhibiting one-sided influence or adaptation, such as fixed decoders, fixed stimulation protocols, or interactions in which only the AI adapts, do not qualify as BHI. The defining feature of BHI is reciprocal, sustained adaptation between biological and artificial intelligence.

\subsection{Living Intelligence}

In BHI, Living Intelligence (LI) is a living neural substrate, typically a dissociated neuronal culture or brain organoid maintained in vitro. These preparations generate spontaneous and stimulus-evoked activity, undergo synaptic reconfiguration, and exhibit high-dimensional spatiotemporal dynamics \cite{mossink2021human}. The BIL provides the recording and stimulation pathways through which LI participates in the closed loop~\cite{kagan2022vitro, cai2023brain}. LI contributes to the distributed computation performed by a BHI system through plasticity and dynamic population activity. Activity-dependent changes in synaptic and cellular state provide a basis for continual, context-dependent adaptation and retain the effects of prior interaction~\cite{khajehnejad2025dynamic}. At the network level, population activity represents information across space and time and transforms inputs non-linearly, properties used in reservoir-style computation~\cite{cai2023brain}. In living neural networks, memory and computation are therefore co-located: synaptic and cellular states both retain the effects of previous activity and shape subsequent processing, so storage and transformation occur within the same physical substrate, unlike von Neumann architectures, which separate memory from processing~\cite{khajehnejad2025dynamic}. These properties may offer computational advantages, but such benefits remain conditional. In vitro networks vary across cultures and over time, and potential advantages such as resilience to perturbations or energy-efficient event-driven processing must be evaluated at the system level; reliable resilience remains to be demonstrated in BHI systems, and any energy advantage must include the costs of culture maintenance, recording, stimulation, and digital control. LI also lacks a directly accessible analogue of gradient-based optimisation, and its behaviour emerges from biochemical and network dynamics that no single computational model fully captures~\cite{lillicrap2020backpropagation,bassett2017network}. A BHI framework must therefore accommodate a computational substrate whose state, response characteristics, and capabilities may change during operation and remain only partially predictable \cite{yaron2025dissociated}.

\subsection{Artificial Intelligence}

In BHI, the AI is the digital computational substrate responsible for the task-relevant functions allocated to it within the distributed computation of the overall BHI system. It receives decoded biological activity together with relevant task or environmental state, and produces outputs such as predictions, actions, intermediate outputs, or stimulation policies. Its computational role depends on the operating mode. It may compete with LI, perform a complementary part of the task, or participate in a computation for which both substrates are required. Unlike a fixed digital controller, the AI must adapt as part of the closed-loop interaction with LI. Biological activity must therefore modify a learning-relevant internal state, such as its model, parameters, or control policy, so that subsequent AI behaviour reflects prior interaction with the living substrate. This allows the AI to combine explicit optimisation, memory, and planning with information emerging from the adaptive dynamics of LI.

\subsection{Orchestrator}

The Orchestrator is the coordination layer responsible for managing interactions and computational responsibility across the system. It assigns computational roles according to the operating mode, schedules exchanges across the different timescales of the two substrates, routes information through the BIL, and integrates their outputs when required by the task. During operation, it tracks the state of the interaction and applies the corresponding coordination policy, determining which biological signals are requested, when AI outputs should influence LI, and when their computational contributions should be integrated. These functions present similarities with established coordination mechanisms in multi-agent systems (\cite{gerkey2004formal}), including allocation, scheduling, routing, state management, and integration, but in BHI they operate across heterogeneous biological and artificial substrates rather than among artificial agents. The BIL implements these exchanges at the signal and stimulation level, including acquisition, encoding/decoding, and safety constraints. The Orchestrator also enforces system-level operating constraints and may delay, modify, or halt an exchange when operating limits are reached. This separation allows the same framework to support adversarial, collaborative, and codependent coordination while preserving AI and LI as the task-computing substrates.

\subsection{Biohybrid Interface Layer}

The BIL is the physical and signal-processing pathway between LI and the digital components. It records extracellular neural activity, converts it into digital features, and translates stimulation commands into precisely timed, charge-balanced pulses \cite{kadan2025electrical}. In neuronal implementations, the BIL is commonly implemented using high-density microelectrode arrays (MEAs), which support parallel extracellular recording and electrical stimulation~\cite{bruno2023neuromorphic,kagan2022vitro}. A BIL must provide sufficient temporal resolution, signal fidelity, bidirectional bandwidth, and stability for the targeted task and adaptation mechanism \cite{hua2025microelectrode}. Existing systems have achieved spike-to-stimulation latency below 10\,ms~\cite{kagan2022vitro}, but this is an implementation result rather than a universal threshold. Sustained BIL operation also depends on maintaining viable cultures under controlled conditions, which vary across in vitro models~\cite{vassal2026neurons}. The BIL constrains what information can pass between the biological and digital domains: its channel count, sampling rate, spatial coverage, latency, noise level, and stimulation precision determine which neural dynamics can be observed and perturbed. These quantities should be reported as interface specifications rather than treated as direct measures of BHI performance. Alternative interface technologies, including patch-clamp electrophysiology, optogenetics, and calcium imaging, offer different combinations of spatial resolution, temporal resolution, invasiveness, and bidirectionality~\cite{wang2025microelectrode,smirnova2023organoid}. MEAs are a practical choice when repeated, multi-site electrical recording and stimulation are both required, but the appropriate interface depends on the biological substrate and task.

\subsection{Embodiment via Cyber-Physical Systems}

Embodiment of BHI allows interactions with CPS, moving from lab-centered applications beyond, but it is optional and is not itself a condition for BHI. We include this interaction towards framework completeness as it has been present in the literature \cite{naughton2026neural}. When a CPS is present, the Orchestrator routes BHI outputs to actuators and incorporates the resulting sensor data to the closed loop. Environmental consequences can influence LI when these data are encoded as stimulation through the BIL, while the AI receives them as task or state information. This embodied pathway can provide AI-to-LI influence when AI-generated actions alter the subsequent sensory input received by LI. Embodiment alone, however, is insufficient for BHI; the system must also satisfy the previously defined requirements for reciprocal, interaction-dependent adaptation. A CPS therefore provides a setting for evaluating co-adaptation through observable behaviour in robotics, navigation, and adaptive control~\cite{ades2024biohybrid,chen2023overview,iida2011soft}.

\section{Modes of Co-Adaptive Operation}

BHI supports multiple coordination strategies between AI and LI, which define its operating modes. We distinguish three: Adversarial, Collaborative, and Codependent. They differ in how computational responsibility is distributed between AI and LI and in how tightly the two substrates are coupled, where each draws on established computational or machine learning paradigms for its design. The following subsections illustrate each mode through a representative use case, highlighting the relationships between framework components, the tasks performed by each component, and the temporal dependencies between them.

\subsection{Adversarial Mode}

\begin{figure}
  \centering
  \includegraphics[width=1\linewidth]{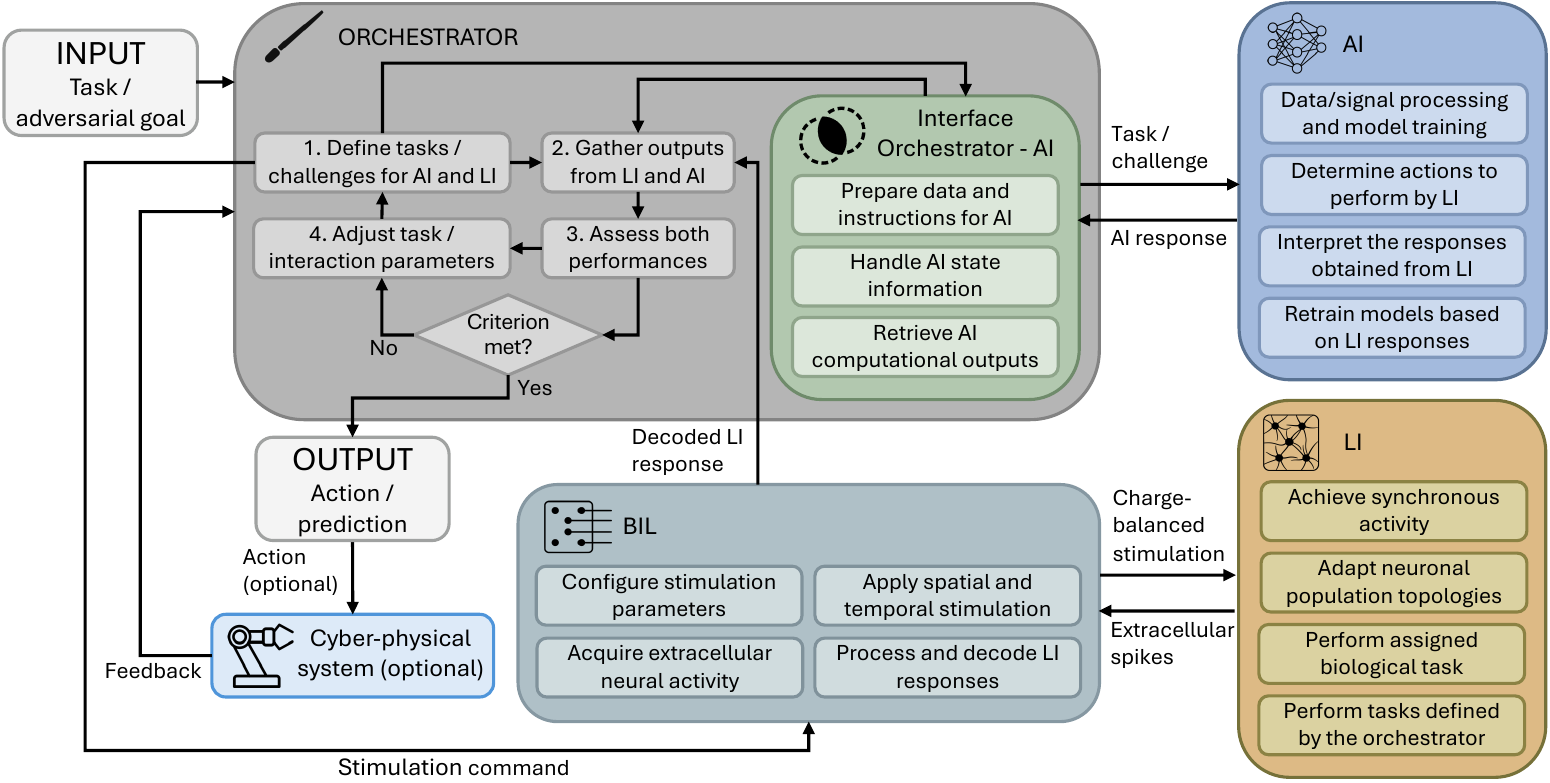}
  \caption{Adversarial mode. The Orchestrator coordinates a competitive interaction between AI and LI, adapting tasks and interaction parameters according to their responses and performance. AI and LI perform the task computation, while the Orchestrator manages the coordination.}
  \label{fig:adversarial}
\end{figure}

In Adversarial Mode, also called the competing mode, the Orchestrator coordinates AI and LI so that they challenge and refine each other through competition. The strategy draws on game theory and adversarial machine learning, where participants improve by competing against opponents (\cite{ao2022eat}). Rather than dividing a task between the substrates, this mode presents competing challenges to AI and LI and uses their responses to determine subsequent interactions (Figure~\ref{fig:adversarial}). The Orchestrator manages this process by coordinating tasks, tracking performance, and adjusting interaction parameters iteratively, while AI and LI remain responsible for task computation and adaptation. A representative use case could be a temporal pattern-learning task in which the AI generates stimulation patterns to challenge the response of LI. The biological substrate adapts to these stimuli, and its recorded responses are then used to update the AI model. As LI becomes better adapted to the patterns, the AI can progressively generate different or more complex stimulation profiles, creating an iterative competition in which each substrate changes the conditions faced by the other. This mode could be useful for studying if reciprocal competition exposes adaptation strategies or response dynamics that would remain unexplored under fixed training conditions. Its main challenges are maintaining stable bidirectional interaction across heterogeneous timescales, controlling biological variability, and ensuring that repeated competitive stimulation remains within physiological and safety constraints.

\subsection{Collaborative Mode}

\begin{figure}
  \centering
  \includegraphics[width=1\linewidth]{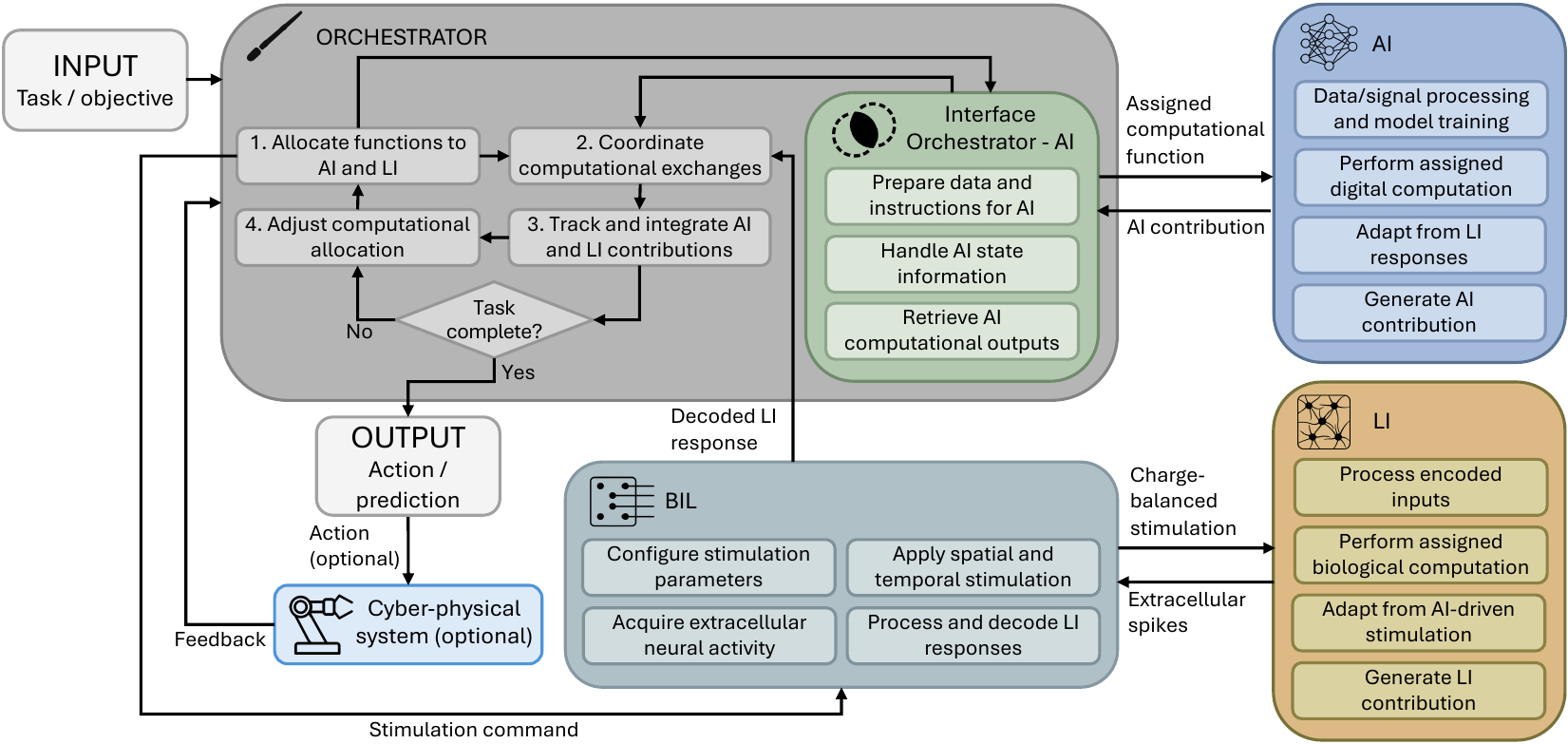}
  \caption{Collaborative mode. The Orchestrator allocates distinct computational functions to AI and LI according to their capabilities and coordinates their interaction so that both substrates contribute to a joint task solution. AI and LI perform complementary task computation, while the Orchestrator manages the division of computational responsibility.}
  \label{fig:collaborative}
\end{figure}

In Collaborative Mode, also called the distributed mode, the Orchestrator allocates computational functions between AI and LI so that they work together rather than compete, each taking the part that suits its strengths (Figure~\ref{fig:collaborative}). LI may contribute adaptive, non-linear exploration or transformation of information, while AI could provide functions such as explicit optimisation, long-range memory, constraint handling, and planning. The Orchestrator coordinates these contributions and their exchange through the BIL so that they jointly contribute to the task solution. Consider a cultured neuronal network controlling a mobile robot navigating an unfamiliar environment with changing obstacles and surface conditions. Through the BIL, sequences of proximity, tactile, and motion signals would be encoded as electrical stimulation, and the resulting neuronal dynamics would provide a non-linear representation of contact state, slippage, and environmental transitions. AI would decode this activity, retain a map and trajectory history, enforce collision and energy constraints, and select the next movement or sensory condition. Each action would alter the robot's subsequent sensory input, producing a closed loop in which LI transforms rapidly changing sensorimotor signals and AI performs explicit state estimation, planning, and constraint handling. Such a configuration could be useful for investigating whether heterogeneous biological and artificial computation can improve exploration, adaptability, or search efficiency in problems where no single computational substrate is best suited to solve the entire problem. Its main challenges are coordinating heterogeneous timescales, maintaining sufficient interface bandwidth, and ensuring that the computational contribution of each substrate can be measured and reproduced.

\subsection{Codependent Mode}

\begin{figure}
  \centering
  \includegraphics[width=\linewidth]{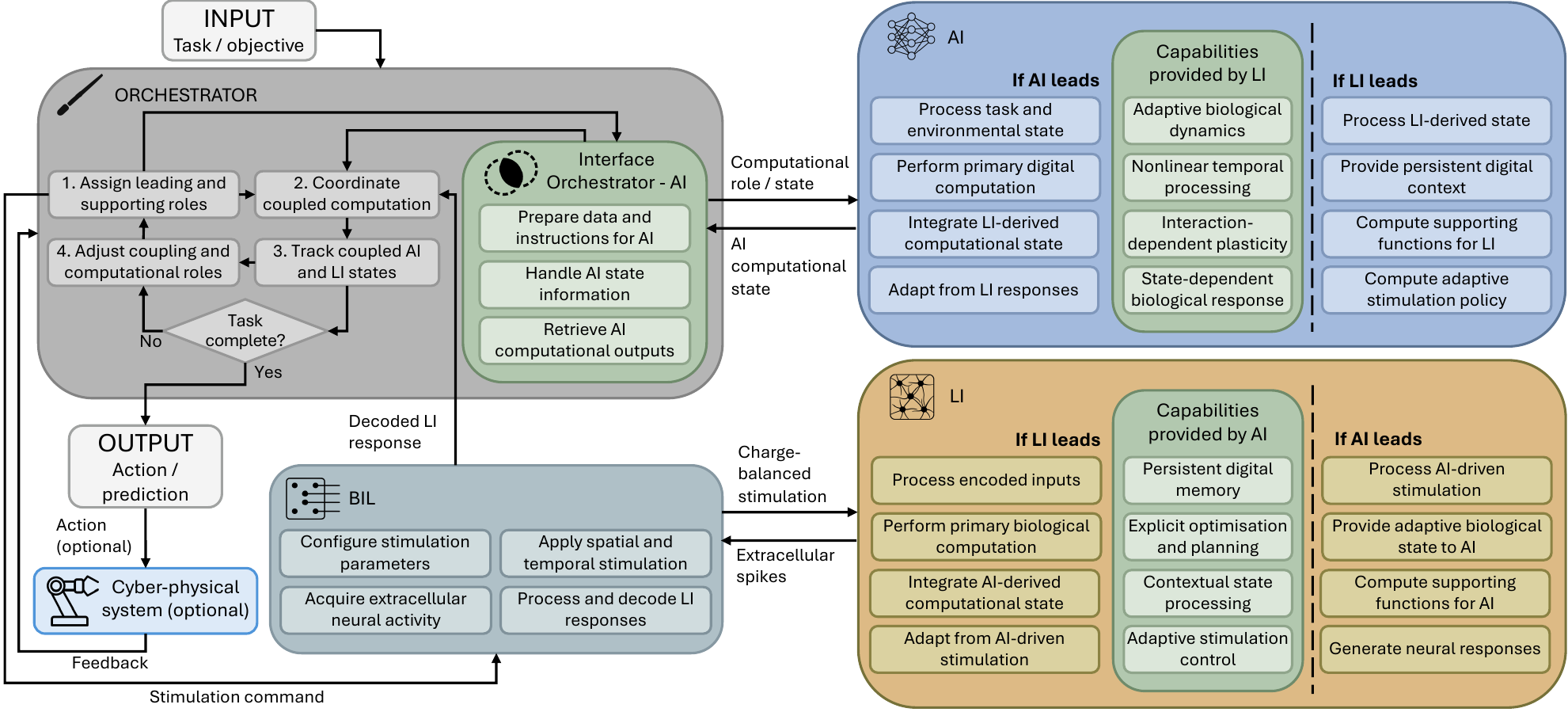}
  \caption{Codependent mode. The Orchestrator couples LI and the AI so that the computation led by one substrate depends on states or capabilities provided by the other. Either AI or LI may assume the leading role, while the other provides supporting computation and capabilities required for task execution. The Orchestrator assigns and adjusts these roles, coordinates the coupled computation, and tracks their interdependent states; AI and LI remain responsible for task computation.}
  \label{fig:codependent}
\end{figure}

In Codependent Mode, the Orchestrator couples AI and LI so closely that one cannot complete the task without the changing state supplied by the other (Figure~\ref{fig:codependent}). Either substrate may lead, but the defining feature is continuous dependence rather than a simple exchange of results. Consider an AI system searching for an effective treatment schedule using a disease organoid—a small laboratory-grown model of a patient’s disease \cite{smirnova2024promise}. The AI would choose which drug to test, at what dose, and in what sequence. The organoid would then provide a living response, such as changes in growth, cell death, or sensitivity to the next treatment. These responses would allow the AI to refine its model and choose the next intervention. Crucially, each decision changes the organoid itself (\cite{huismans2025impact}): exposure to one drug may make the disease model more resistant or more sensitive to another. The AI is therefore not analysing a fixed biological dataset since its decisions alter the system that generates the next data. Without the organoid, the AI loses access to the evolving treatment response; without the AI, the organoid can respond to drugs but cannot compare treatment strategies or decide what should be tested next. Replacing the living responses with stored or shuffled measurements would also break the computation because the next response must depend on the previous intervention. This distinguishes Codependent Mode from Collaborative Mode, where AI and LI perform separate functions and their outputs are subsequently combined. The main engineering challenges are maintaining reliable communication between the digital and biological components, coping with their different operating speeds, detecting changes caused by biological drift rather than treatment, and recovering safely when the interface fails. Evaluation must therefore measure not only whether the system finds an effective treatment sequence, but also whether its performance depends on the intact two-way loop.

\section{Advantages and Distinctions of BHI}

The contribution of BHI is not the introduction of a new biological substrate, interface technology, or machine learning method. Rather, BHI proposes a framework for organising computation across adaptive biological and artificial substrates that interact within the same closed loop. Existing paradigms address important components of this problem. \textit{Synthetic Biological Intelligence}~\cite{kagan2025two} uses living systems as computational substrates but does not generally require reciprocal adaptation with an artificial learning system. \textit{Neuromorphic computing}~\cite{chiappalone2026advancing,bruno2023neuromorphic} reproduces biological computational principles in silicon and therefore operates without living tissue. \textit{Organoid intelligence}~\cite{smirnova2023organoid} investigates the computational capabilities of organoids and neuronal cultures but does not require an adaptive AI component participating in closed-loop co-adaptation. \textit{Neural decoding and conventional neurotechnologies} extract information from biological activity, but this information need not update a co-adapting artificial partner. \textit{None of these paradigms explicitly defines how computational responsibility should be distributed, coordinated, and evaluated when both biological and artificial substrates adapt during operation.} Table~\ref{tab:bhi_comparison} summarises these distinctions.

\begin{table}[t]
\centering
\footnotesize
\caption{Defining characteristics of BHI relative to related paradigms.}
\label{tab:bhi_comparison}

\begin{tabularx}{\linewidth}{>{\raggedright\arraybackslash}Xcccc}
\toprule
\textbf{Paradigm} &
\begin{tabular}{@{}l@{}}
\textbf{Living} \\
\textbf{substrate}
\end{tabular} &
\begin{tabular}{@{}l@{}}
\textbf{Adaptive} \\
\textbf{AI}
\end{tabular} &
\begin{tabular}{@{}l@{}}
\textbf{Reciprocal} \\
\textbf{adaptation}
\end{tabular} &
\begin{tabular}{@{}l@{}}
\textbf{Explicit computational} \\
\textbf{allocation}
\end{tabular} \\
\midrule
Neural decoding / Neurotech & Yes & Optional & No & No \\
\hline
Synthetic Biological Intelligence & Yes & Typically no & No & Limited \\
\hline
Neuromorphic computing & No & Yes & No & No \\
\hline
Organoid intelligence & Yes & Optional & Not required & Limited \\
\hline
BHI & Yes & Yes & Required & Required \\
\bottomrule
\end{tabularx}
\end{table}

These distinctions also affect how BHI systems should be evaluated. Task performance alone is insufficient: evaluation should assess the adaptation in both substrates, identify their respective computational contributions, and report the relevant BIL specifications under which their interaction occurs. Relevant BIL specifications, including latency, bandwidth, and stimulation constraints, should therefore be considered alongside changes in AI and LI during operation to determine whether observed improvements emerge from distributed biological–artificial computation rather than from a single dominant component.

The advantages of combining biological and artificial substrates are conditional rather than automatic. Biological neural networks operate through sparse, event-driven signalling and integrate memory and computation within the same physical substrate \cite{zhou2025exploiting}. In principle, these properties may support continual adaptation, rapid learning from limited experience, and energy-efficient information processing. However, whether such advantages translate into system-level gains depends on the complete computational budget, including culture maintenance, recording, stimulation, control, and interface overhead. Likewise, reports of high sample efficiency in biological neural systems~\cite{khajehnejad2025dynamic} remain task-dependent and require validation across a broader range of benchmarks. The relevant comparison is therefore not between biological and artificial computation in isolation, but between alternative allocations of computation across the complete BHI system.

Beyond these architectural questions, the main barriers to practical BHI systems are currently engineering and experimental. Variability between biological preparations, limited long-term stability, restricted interface bandwidth, and reproducibility across laboratories remain the dominant constraints on practical deployment~\cite{vassal2026neurons}. Consequently, the value of BHI should not be assessed by the capabilities of the biological substrate alone, but by whether distributed biological--artificial computation produces measurable performance, adaptability, or efficiency gains beyond what either substrate can achieve independently.

\section{Roadmap for Reproducible BHI Development}

\textbf{Near-Term Horizon:} Current biological substrates suffer from high trial-to-trial variability and limited operational shelf-life~\cite{vassal2026neurons}. The primary objective for the near-term is to stabilise LI against three targets. The first is closed-loop latency below 10\,ms: this matches the spike-timing-dependent plasticity window (\cite{andrade2023timing}) and is the minimum at which the Orchestrator can drive biological learning in real time~\cite{kagan2022vitro}. The second proposed target is continuous closed-loop task operation for at least 24,h, with predefined limits on task-performance drift, evoked-response stability, and culture viability. This is an engineering milestone rather than an established biological threshold. Current wetware platforms support 24/7 electrophysiological monitoring and organoid operation exceeding 100 days, while chronic interfaces have recorded organoid activity for up to 120 days \cite{jordan2024open,yang2024kirigami}. However, adaptive task demonstrations remain substantially shorter, including 20-min gameplay sessions \cite{khajehnejad2025dynamic}. The unresolved challenge is therefore sustained adaptive computation, not culture survival or recording alone.The third is reproducibility: as the community must release raw electrophysiology data and stimulation protocols for standard control tasks and report results across multiple distinct cultures, to increase lab-lab comparisons towards computing guarantees across different settings.

\textbf{Mid-Term Horizon:} Once individual platforms achieve stability, the focus must shift to community-wide standardisation. The critical deliverable in this phase is a set of BHI benchmarks---analogous to ImageNet in digital AI---that quantify biological-artificial computational performance in bits-per-joule, learning efficiency, and computational contribution of each substrate. Three milestones define this phase. Nonlinear function-approximation tasks must be validated across multiple laboratories, so that a reported capability is known to be a property of the substrate rather than of one local setup. The field needs a formal culture specification---listing input/output limits, frequency response, and metabolic requirements, in effect a datasheet for a biological component---so that results can be compared on common terms. And safety and data-governance checklists must be adopted across major research sites, which lowers the barrier for new groups to enter and keeps practice consistent as the field grows.

\textbf{Applications:} Potential application areas arise where distributed biological-artificial computation may offer advantages over either substrate alone. In robotics, BHI embeds living neural cultures within control frameworks, allowing online learning, context-sensitive decision-making, and experience-dependent plasticity that can increase autonomy, behavioural refinement, and limited self-repair relative to silicon-only controllers~\cite{chen2023overview,iida2011soft}; this is the most direct instantiation of co-adaptation, since the culture adapts to the consequences of action while the AI adapts to the culture. In neuromorphic computing, biological substrates can inform circuits that exploit neural spatiotemporal dynamics for low-power, high-throughput processing. In personalised medicine, biohybrid implants and prosthetics could co-adapt with an individual's physiology over time, and BHI platforms configured to reproduce patient-specific phenotypes could support controlled drug-response testing and therapeutic optimisation. In environmental monitoring, bio-integrated sensors could deliver context-aware, real-time detection with higher specificity under complex conditions, with distributed sensor--actuator deployment.

\section{Conclusion}
BHI envisions living intelligence and artificial intelligence as co-adaptive computational substrates whose task-relevant computations are distributed and coordinated within a shared closed loop. Through the BIL and Orchestrator, the framework makes computational responsibility explicit and distinguishes systems that merely exchange biological and digital signals from those in which both substrates contribute to task computation and adapt through their interaction. The proposed modes further describe how this relationship can be organised through competition, complementary computation, or computational interdependence. In this way, BHI provides a common framework for describing and comparing systems that are otherwise addressed separately across biological and artificial computing paradigms.

The decisive question is now whether distributed biological–artificial computation can produce reproducible and measurable advantages over either substrate operating independently. Addressing this question requires stable long-term operation, reproducible experimental protocols, well-characterised interfaces, and benchmarks that quantify not only task performance but also adaptation and the computational contribution of each substrate. Whether the resulting advantages lie in adaptability, learning efficiency, energy efficiency, or other properties remains an empirical question. Establishing such evidence will determine whether BHI develops from a framework for organising co-adaptive systems into a robust computational paradigm.

%% \begin{acks}
%% Add exact funding sources and acknowledgements here before submission.
%% Left commented out deliberately: a placeholder acknowledgements block
%% signals an unfinished manuscript to reviewers.
%% \end{acks}

\bibliographystyle{unsrtnat}
\bibliography{references}

\end{document}